\documentclass{article} 
\usepackage[preprint]{colm2025_conference}   

\usepackage{amsmath,amssymb,amsfonts}
\usepackage{algorithmic}
\usepackage{microtype}
\usepackage{hyperref}
\usepackage{url}
\usepackage{booktabs}
\usepackage{multirow}
\usepackage{longtable}

\usepackage{graphicx}
\usepackage{textcomp}
\usepackage{xcolor}

\usepackage{lineno}
\usepackage{algorithm}

\definecolor{darkblue}{rgb}{0, 0, 0.5}
\hypersetup{colorlinks=true, citecolor=darkblue, linkcolor=darkblue, urlcolor=darkblue}

\title{Learning to Rewrite Prompts for Bootstrapping LLMs on Downstream Tasks}

\author{Qinhao Zhou, Xiang Xiang\thanks{Correspondence to \url{xex@hust.edu.cn}. Project done at HUST AI \& Visual Learning (HAIV) Lab.}, Kun He \& John E. Hopcroft \\
Hopcroft Center on Computing Science \\
Huazhong University of Science and Technology (HUST), China
}

\begin{document}

\ifcolmsubmission
\linenumbers
\fi

\maketitle

\begin{abstract}

In recent years, the growing interest in Large Language Models (LLMs) has significantly advanced prompt engineering, transitioning from manual design to model-based optimization. Prompts for LLMs generally comprise two components: the \textit{instruction}, which defines the task or objective, and the \textit{input}, which is tailored to the instruction type. In natural language generation (NLG) tasks such as machine translation, the \textit{input} component is particularly critical, while the \textit{instruction} component tends to be concise. Existing prompt engineering methods primarily focus on optimizing the \textit{instruction} component for general tasks, often requiring large-parameter LLMs as auxiliary tools. However, these approaches exhibit limited applicability for tasks like machine translation, where the \textit{input} component plays a more pivotal role. To address this limitation, this paper introduces a novel prompt optimization method specifically designed for machine translation tasks. The proposed approach employs a small-parameter model trained using a back-translation-based strategy, significantly reducing training overhead for single-task optimization while delivering highly effective performance. With certain adaptations, this method can also be extended to other downstream tasks.
\end{abstract}



\section{Introduction}

LLMs \cite{brown2020language, touvron2023llama, chowdhery2022palm, bai2023qwen, du2021glm} have achieved groundbreaking progress across a myriad of pattern recognition tasks. These LLMs, such as GPT-3.5 \cite{gpt-3.5}, typically use manually crafted or predefined prompt templates as directives to guide the model in accomplishing various tasks. Compared to task-specific models, LLMs possess broader knowledge and enhanced expressive capabilities. In traditional NLG and NLU tasks, LLMs can outperform smaller models designed for specific tasks. Furthermore, when subjected to downstream task-specific fine-tuning, LLMs achieve even more competitive results \cite{Alves2024TowerAO}.

While LLMs have demonstrated remarkable performance, researchers \cite{qin2021learning,liu2021gpt} have found that prompts play a critical role in enabling LLMs to accomplish various downstream tasks. Additionally, LLMs exhibit significant sensitivity to prompts, where even minor modifications can lead to entirely different outputs. To enhance the performance of LLMs on downstream tasks, a significant body of work has proposed various prompt engineering methods. For example, soft prompts \cite{qin2021learning} convert discrete prompt words into continuous vectors, enabling end-to-end training. Prefix Tuning \cite{li2021prefix} inserts a series of continuous task-specific prefixes at the beginning of the input, then fine-tunes these prefixes while keeping the other parameters frozen. Furthermore, APO \cite{pryzant2023automatic} optimizes the prompt by using the discrete feedback of the LLMs as gradient updates. RLPrompt \cite{deng2022rlprompt}  employs reinforcement learning to conduct a directionless Monte Carlo search in the semantic space. These methods focus on optimizing the entire prompt for general downstream tasks, achieving highly effective results.
\begin{figure}[t]
  \centering 
  \includegraphics[scale=0.4]{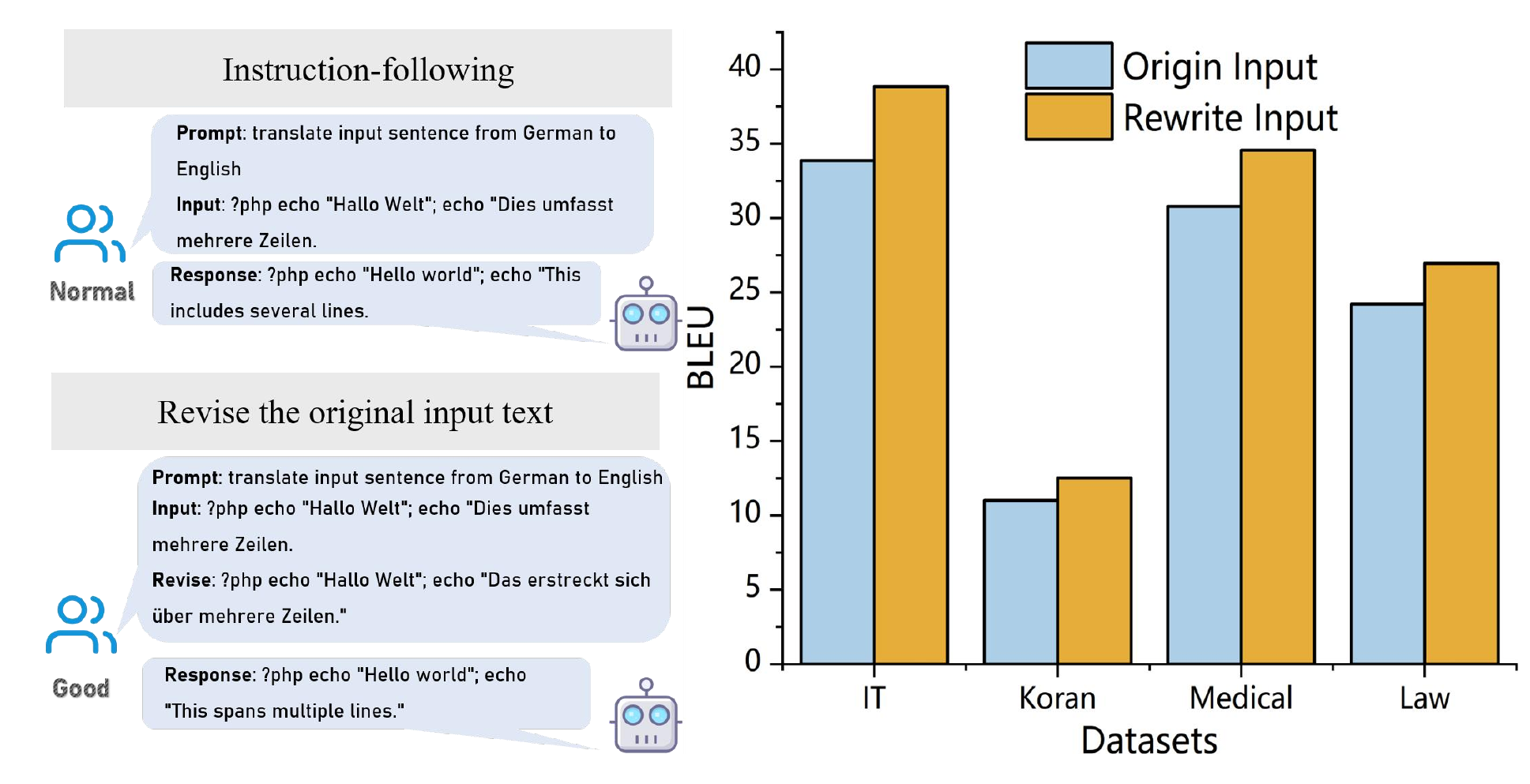}
  \centering
  \caption{ 
  The example on the left illustrates the sensitivity of LLMs to input data. In the context of the same translation task, providing more detailed input (bottom) leads to better results. The right figure shows the performance improvement of the LLMs on various translation datasets after rewriting.
  }
  \label{instruction-following}
  \vspace{-5mm}
\end{figure}

In the context of using LLMs, prompts typically consist of two components \cite{wei2021finetuned}: \textit{instruction} and \textit{input}. These two components are usually concatenated and fed into the LLMs. The aforementioned approaches primarily optimize the $instruction$  to achieve improved results. However, in specific NLG tasks like machine translation, the model is equally sensitive to the $input$. As depicted in Figure~\ref{instruction-following} left, in the task of translating German to English, LLMs provide better responses when the original German text is modified while preserving the original meaning. In tasks such as machine translation, summarization and abstraction, the \textit{instruction} component is typically short and may not require significant modifications. However, the \textit{input} component plays a major role in the input tokens in these tasks. Therefore, keeping prompt fixed and optimizing the \textit{input} part is crucial for enhancing LLMs performance in these tasks.

Inspired by Augmented Language Models (ALMs) \cite{mialon2023augmented}, we propose the Rewriting Original Inputs (ROI) strategy, which aims to optimize the critical \textit{input} component before feeding it into LLMs. This strategy leverages either a smaller parameter model or the LLM itself to reformulate the original input, aligning it more closely with the model's inherent preferences. We first employ the LLM itself to optimize \textit{input} for various tasks. We find that for shorter inputs in NLG tasks, the LLMs produces better reformulations. However, for longer inputs, the lack of fine-tuning for rewriting tasks often leads to hallucinations or alterations of the original meaning. To address this issue, we fine-tune a rewriting model specifically for optimizing the \textit{input}. Since the rewriting model is focused solely on this task, we can use a smaller parameter model, thereby reducing the fine-tuning overhead. We draw inspiration from back translation in machine translation and use this way to construct rewriting data and train the rewriting model. As shown in Fig.~\ref{instruction-following} right, on several translation datasets, the results of LLMs show varying degrees of improvement when the original input data is rewritten while keeping the original meaning. For NLU tasks such as sentiment analysis and syntactic parsing, the \textit{input} part is relatively short, and in tasks like causal reasoning, the \textit{input} is often divided into segments. Consequently, using smaller models for rewriting yields unsatisfactory results. Therefore, for NLU tasks, we utilize the LLM itself to reformulate the \textit{input} without altering the content of the instructions. 

Additionally, recognizing that rewriting is not universally effective, we introduce a filtering module that employs text similarity metrics to assess the quality of rewritten text. This module is designed to identify and eliminate instances of hallucinations and noise, thereby ensuring the integrity and reliability of the rewritten content. Specifically, the filtering mechanism evaluates the semantic consistency between the rewritten text and the original input, discarding outputs that deviate significantly from the intended meaning. For texts that remain noisy or inconsistent after multiple rewriting attempts, we revert to the original \textit{input} to maintain the fidelity of the data.

Regarding the contributions of this paper, we observe that existing prompt engineering methods yield limited benefits for tasks where the \textit{input} component plays a predominant role. Building upon this observation, we then introduce the Rewrite Original Input (ROI) module, coupled with a filtering algorithm, to boostrap the performance of LLMs on these downstream tasks. 
In this method, there is no need to train any parameters in the LLMs and the framework is applicable to a wide range of different LLMs.
The experimental results on both NLU and NLG tasks verify that the ROI module effectively transforms ambiguous data into more precise and explicit input prompts. Compared to the original input, our ROI method reaches consistent and notable performance improvements across all tasks.

\vspace{-1mm}

\section{Related Work}

\vspace{-3mm}

\textbf{Augmenting LLMs with Prompt Tuning.} The utilization of a shared model across tasks has significantly propelled the application and development of LLMs. However, the reliance on textual prompts requires manual design, and even with carefully crafted prompts, their performance still falls short compared to model fine-tuning. As a result, current work primarily aims to enhance the performance of LLMs through differentiable tuning of prompts. \cite{lester2021power} and \cite{li2021prefix} propose a method called prefix tuning to adjust soft prompts for tuning frozen models. The tokens of soft prompts are learnable vectors, and they append the soft prompt vectors at the beginning of the input text, inputting the combined sequence into the model, thus realizing end-to-end training on the training set. Similarly, P-Tuning \cite{liu2021gpt, liu2021p} adds an encoder module in front of LLMs to fine-tune prompts at the embedding level, which is more flexible compared to prefix tuning. In addition, APE \cite{zhou2022large} and RLPrompt \cite{deng2022rlprompt} incorporate reinforcement learning into prompt optimization. They design scoring functions in response to model feedback and make discrete-level corrections to prompts.

\textbf{Augmenting LLMs without Training.} Training LLMs from scratch poses a significant challenge for researchers due to their massive parameter size and the need for extensive pre-training data. LLMs exhibit excellent context-learning capabilities, allowing the completion of specific tasks through contextual prompts, known as in-context learning (ICL) \cite{dong2022survey}. Unlike supervised learning, in-context learning does not require parameter updates but directly uses LLMs for prediction. LLMs can understand given demonstrations and make accurate predictions. The performance of ICL heavily depends on the nature of demonstrations, including both their format and sequence. KATE \cite{liu2021makes} indicates that the selection of nearest-neighbor samples as context instances can significantly enhance the performance of LLMs. Additionally, \cite{gonen2022demystifying} proposes selecting instances with low perplexity, while \cite{rubin2021learning} puts forth a two-stage, retrieval-based method for demonstration selection. To handle specific inputs, an unsupervised retriever is first constructed to identify examples similar to candidate instances, following this, a supervised retriever selects appropriate demonstrations among these candidates.

\textbf{Rewriting Techniques Applied in LLMs.} The strategy of rewriting is widely applicable across various domains. \cite{Kong2024PRewritePR} propose a rewriting framework utilizing reinforcement learning, achieving impressive results. Learning2Rewrite \cite{Hao2024LearningTR} proposes a comprehensive framework for detecting AI-generated text. By leveraging the differences in how LLMs rewrite human-authored text versus AI-generated text, it achieves robust performance in unseen domains. RewriteLM \cite{shu2024rewritelm} enhances the performance of LLMs in cross-sentence rewriting tasks through instruction tuning and reinforcement learning strategies, positioning it as an effective rewriting model. APOHF \cite{lin2024prompt} is a method that optimizes prompts using human feedback, offering greater alignment with real-world applications and achieving impressive results through rapid iteration.


\section{Methodology}
We investigate the performance of 7B-13B parameter LLMs on tasks such as machine translation and summarization, revealing that these models still exhibit performance gaps compared to specialized models in certain domains. Fine-tuning such LLMs risks compromising their general knowledge while incurring significant computational costs. To address this, we employ smaller traditional language models to perform dedicated input rewriting tasks. By replacing original inputs with rewritten versions that better align with LLMs' preferences, we facilitate improved generation quality without modifying the LLMs themselves.

\subsection{Rewriting Original Input with Small Models}
We first introduce our rewriting module in detail. For clarity in our discussion, we formally define the prompt \( P \) input to LLMs as consisting of two distinct components:\[P = (I, X).\] Where \( I \) represents the \textit{instruction part} specifying the task requirements. \( X \) denotes the \textit{input part} containing the content to be processed. Previous studies have demonstrated that the LLMs are highly sensitive to the instruction $I$, highlighting that even slight modifications can result in significant variations in the outputs of the model. As a consequence, a considerable amount of research has emerged that aims to optimize the design of instruction prompts. However, for traditional NLU and NLG tasks such as translation and sentiment analysis, the instruction $I$ is relatively fixed, the benefits of optimizing this aspect are limited. Instead, the input $X$ emerges as more crucial.

\begin{figure*}[t]
  \centering 
  \includegraphics[scale=0.45]{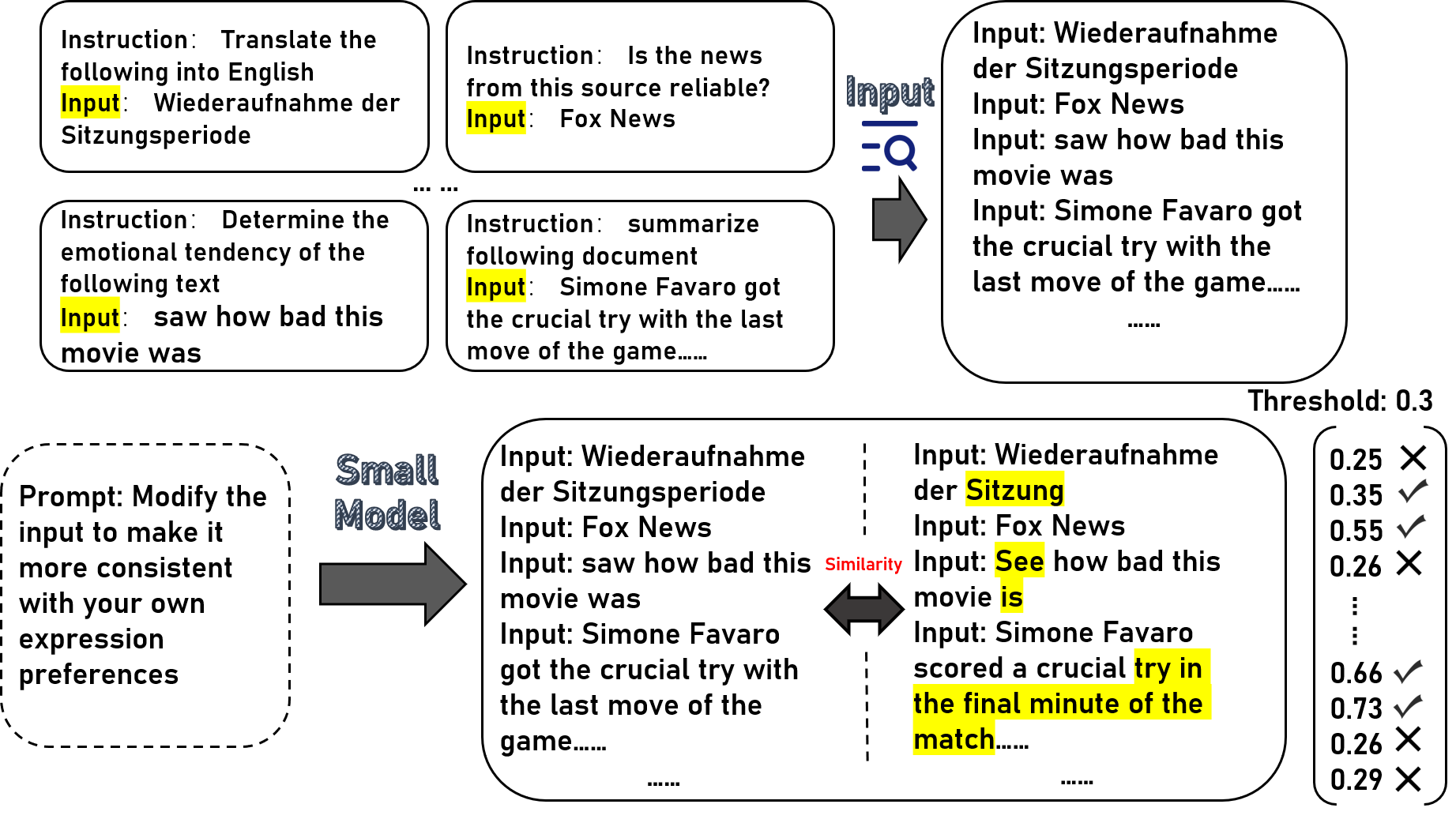}
  \caption{ 
  The pipeline of our proposed method for boostrapping Large Language Models. 1) The raw data is first input to the LLMs to construct the rewriting data.  The generated rewritten data is then used to train the rewriting model. 2) The filtering process is applied to retain only the rewritten data that demonstrates improved performance, while the remaining data continues to utilize the original data. 3) During testing, the original data is first input to the rewriting model to obtain rewritten sentences. These rewritten sentences are subsequently input to the LLMs to generate the final results.
}
  \label{pipeline}
  \vspace{-5mm}
\end{figure*}

We observe that input sentences expressing the same meaning may elicit different responses from the LLMs under the same instruction template. In other words, LLMs are also sensitive to input components. In real-world scenarios, LLMs face the challenge of handling diverse user writing styles and preferences, requiring them to produce coherent and sensible outputs for these varied expressions. To this end, we propose to modify the input data before it is processed by the LLMs, transforming it into the most accessible for the model. We introduce a rewriting module that operates on the input data. This can be expressed by the following equation:

\vspace{-3mm}
\begin{equation}
    y_j = argmax_{y_j \in V}P_M(y_j|R(X),y_{<j}).
    \label{eq1}
\end{equation}

Where $V$ denotes the vocabulary, $y_i$ represents the next token, $X_{Prompt}$ indicates the initial input, and $R(\cdot)$ signfies the rewriting of the initial input. We hypothesize that LLMs have their own preferences regarding the data they process, which may diverge from conventional human expression patterns. Therefore, we design a process where the original input data is rewritten using either a language model with fewer parameters or the LLMs itself. Specifically, inspired by the technique of back translation in machine translation, we utilize LLMs to write back the training set output as input and re-word the rewritten data with the original input of the training set. We then use this train set to fine-tune a language model with fewer parameters and we call this model a rewriting model. The rewriting model learns the preferences of the LLMs towards input data. When new test data is available, we first input the \textit{input} component to the rewriting model and then pass the rewritten result to LLMs for further processing. Figure~\ref{pipeline} shows the overall process.

We give an example of our rewriting method. For the machine translation task from German to English,  we first back-translate the training data from English to German and combine it with the original German input to form the rewritten data. This data is used to train the rewriting model and the rewriting model is a compact model designed solely for the rewriting task. During testing, the input is first rewritten by the rewriting model to align with the model's preferences, and then it is fed into the LLMs.

For judgment tasks involving grammar, sentiment, etc., there is no one-to-one correspondence between input and output. In this scenario, where it is not possible to construct a rewriting dataset, we leverage the capabilities of the LLMs themselves to perform rewriting, so that the input component adapts to the preferences of LLMs themselves. For instance, in SST task, when dealing with data exhibiting a positive emotional tendency, the prompt is formulated as "Modify the input sentence to enhance its positive emotional tendency without altering the original meaning." Conversely, when the data exhibits a negative emotional tendency, the prompt is adjusted to "Modify the input sentence to amplify its negative emotional tendency without changing the original meaning."

 Our method is not applicable to tasks involving reasoning, planning, and other abilities. In these tasks, the input component is relatively fixed, and prompts need to focus more on activating the reasoning capabilities of the LLMs, so methods such as Chain of Thoughts \cite{wei2022chain} are more suitable. Furthermore, we have found that not all data receive positive benefits from rewriting. Due to the unstable output of LLMs, they sometimes produce so-called hallucinations. Therefore, it is necessary to filter and select the data after rewriting.

\begin{algorithm}[h]
\caption{Filtering Algorithm}
\renewcommand{\algorithmicrequire}{\textbf{Input:}}
\renewcommand{\algorithmicensure}{\textbf{Output:}}
    \begin{algorithmic}[1]
        \REQUIRE Rewrite dataset $\mathcal{R}=\{ \emptyset \}$, Rwrite function $\mathcal{F}$, Original dataset $\mathcal{D}=\{{(x_1, y_1), (x_2, y_2),...,(x_n, y_n)}\}$
        \ENSURE Rewrite dataset $\mathcal{R}$
        \STATE {Rewrite the original statement and qualify it}
        \FOR{$(x_i, y_i) \in D$}
            \STATE $r_i = F_{task}(x_i)$ or $r_i = F_{task}(x_i, y_i)$;
            \STATE $sim_{score} = metric(r_i, x_i)$;
            \IF{$sim_{score} < \gamma$}
                \STATE $R_i = x_i$;
            \ELSE
                \STATE $R_i = r_i$;
            \ENDIF
        \ENDFOR
        \STATE \textbf{return} $R$;
    \end{algorithmic}
\label{algorithm 1}
\end{algorithm}

\subsection{Filtering Noise with Similarity Computation}

During the rewriting process, it is inevitable that some noise data will be generated, and not all rewrites are beneficial. To address this, we introduce a filtering mechanism that follows the rewriting model. Specifically, for different tasks, we calculate similarity using pertinent evaluation metrics and set thresholds for filtering. For instance, in a translation task, we can use word-level edit distance to calculate the similarity between the original text and the rewritten sentences. When the similarity between the rewritten sentences and the original text is low, it might be because LLMs have outputted hallucinations, or that extensive rewriting increases the training difficulty for the rewriting model. Therefore, we replace them with the original text, preserving only the rewritten data that have a small degree of change and are effective. We conduct filtering texts using three metrics: RougeL, BLEU, and word-level edit distance. We utilize the ROUGE-L metric to calculate the similarity. Only when the ROUGE-L score between the original and the rewritten sentence surpasses a certain threshold, we add it to the rewritten dataset. Furthermore, as rewriting is analogous to a language translation task, we use BLEU as another metric to evaluate similarity. Rewriting often involves rearranging word orders, deleting inappropriate words, adding new terms, etc., which is directly related to the concept of edit distance. Therefore, we also adopt edit distance as a similarity measure. Suppose $s$ and $t$ are two sentences and the relevant formula is as follows:

\begin{equation}
    ER = \frac{D(s,t)}{|s_{len}|}.
\end{equation}
Where $|s_{len}|$ is the length of sentence $s$ and $D(s,t)$ is the edit distance function, which is computed as follows:

\begin{equation}
D(i,j) = 
\begin{cases}
\max(i,j) & \text{if } \min(i,j) = 0, \\
D(i-1,j-1) & \text{if } s[i] = t[j], \\
1 + \min \big( D(i-1,j), D(i,j-1), D(i-1,j-1) \big) & \text{otherwise.}
\end{cases}
\end{equation}

Where $i$ and $j$ are the index of the two sentences. The filtering process is detailed in Algorithm \ref{algorithm 1}.

\section{Experiments}
\subsection{Datasets and Setup}

We conduct experiments on both NLG and NLU tasks to investigate the impact of our method on various downstream tasks. For NLG tasks, we conduct experiments on machine translation and summarization tasks. For translation task, we utilize four different domain-specific German-to-English translation datasets: IT, Medical, Koran, and Law. Tab.~\ref{dataset} shows the divisions of four de-en translation datasets that we used. Both the IT and Medical domains have over $20000$ training samples, while Laws has over $46000$. The Koran dataset is relatively smaller with only $17982$ training data points. For summarization task, we conduct experiments on Xsum \cite{Narayan2018DontGM} dataset. It consists of BBC news articles paired with single-sentence summaries, which is widely used in summarization tasks. For the NLU task, we chose the GLUE benchmark \cite{wang2018glue}, consisting of nine different NLU tasks. It combines instruction prompts and data inputs. Therefore, we modify and optimize the entire input. We report the BLEU score, Edit Rate and RougeL for NLG tasks and the average accuracy and average F1 score for NLU tasks.

\begin{table}[htbp]
\centering

\scalebox{1.20}{
\begin{tabular}{ccccc}\toprule
Dataset & IT & Medical & Koran & Laws \\ \hline
Train & 222, 927 & 248, 009 & 17, 982 & 467, 309 \\
Test & 2000 & 2000 & 2000 & 2000 \\ 
\bottomrule
\end{tabular}
}
\caption{Statistics of dataset in different domains for translation task.}
\vspace{-3mm}
\label{dataset}
\end{table}

\subsection{Implementation Details}

We conduct experiments on different versions of the Alpaca \cite{taori2023alpaca} model with varying parameter sizes. The original alpaca model is based on the LLaMA model and is fine-tuned with $52k$ instructions. For the rewrite models, we select models with fewer parameters, such as mBart \cite{liu2020multilingual} and mT5 \cite{xue2020mt5}. For different versions of the alpaca model, we set the temperature coefficient to $0.1$ and the number beams to $4$. The initial learning rate is set to $2e^{-5}$, the batch size is $4$, and the dropout is $0.3$. 

\subsection{Main Results}

The experimental results of using ROI in NLG tasks are presented in Table~\ref{translation results}. In NLG task, the \textit{input} part plays a crucial role in determining the quality of the output and it can be seen that optimizing input significantly enhances the model's output evaluation performance. In Medical dataset, compared to using the original inputs, the ROI method achieves a BLEU score increase of $2.9$ and an improvement of $1\%$ in Edit Rate. In Xsum summarization task, the RougeL increased by $0.28$ after rewriting input. We observe that our method shows limited effectiveness in some domains. In the Koran domain, our method shows a positive enhancement of $0.14$ in BLEU and a slight decrease in edit distance. In the IT domain, our method has an improvement of $0.31$ in BLUE and a slight decrease in edit distance. We attribute this to the varying difficulty of data across different domains and the extent of rewriting required. For particularly challenging data, such as the Koran, rewriting the original text is relatively difficult. Conversely, for data that requires minimal rewriting, such as IT texts, the impact of rewriting is limited.


\begin{table}[h]
\centering

\scalebox{1}{
\begin{tabular}{ccccc} \toprule
                         &         & BLEU                 & Edit Rate            & RougeL \\ \hline 
\multirow{2}{*}{IT}      & origin  & 27.75                & 0.62                 &   -     \\
                         & rewrite & 28.06                & 0.61                 &   -     \\
\multirow{2}{*}{Koran}   & origin  & 12.37                & 0.57                 &   -     \\
                         & rewrite & 12.51                & 0.56                 &   -     \\
\multirow{2}{*}{Medical} & origin  & 31.67                & 0.63                 &   -     \\
                         & rewrite & 34.57                & 0.64                 &   -     \\
\multirow{2}{*}{Law}     & origin  & 24.21                & 0.63                 &   -     \\
                         & rewrite & 26.94                & 0.63                 &   -     \\ \hline
\multirow{2}{*}{Xsum}    & origin  & \multicolumn{1}{l}{-} & \multicolumn{1}{l}{-} & 0.13   \\
                         & rewrite & \multicolumn{1}{l}{-} & \multicolumn{1}{l}{-} & 0.41   \\
\bottomrule
\end{tabular}
}
\caption{
    The experimental results for NLG tasks.
}
\label{translation results}
\end{table}

\begin{table}[h]
\centering

\scalebox{1.20}{
\begin{tabular}{cccc} \toprule
                                  &              & Accuracy & F1 Score \\ \hline 
\multirow{2}{*}{llama-7b-hf}      & origin & 57.87    & 52.93    \\
                                  & rewrite      & 59.02    & 57.44    \\
\multirow{2}{*}{flan-alpaca-gpt4} & origin & 56.43    & 50.39    \\
                                  & rewrite      & 57.07    & 50.79     \\
                                 \bottomrule
\end{tabular}
}

\caption{
    The experimental results for NLU tasks.
}
\label{GLUE results}
\end{table}

Table~\ref{GLUE results} presents the average accuracy and F1 score of ROI and original inputs on various NLU tasks. We use the same set of training parameters for all tasks and the table show the average performance across all NLU tasks we test. For each model and each evaluation metric, the original and our rewritten results are listed separately. The results demonstrate that ROI can be applied to various downstream tasks. Both accuracy and F1 score show improvements with our method compared to using the original input. Additionally, our approach achieves favorable results across different LLMs. In the Llama-7B-HF model, rewriting yields a $1.16\%$ increase in accuracy and a $4.51$ improvement in F1 score. Additionally, in the Flan-Alpaca-GPT-4 model, rewriting results in a $0.6\%$ increase in accuracy and a $0.4$ improvement in F1 score.

\begin{figure}[hbp]
  \centering 
    \includegraphics[width=0.90\textwidth]{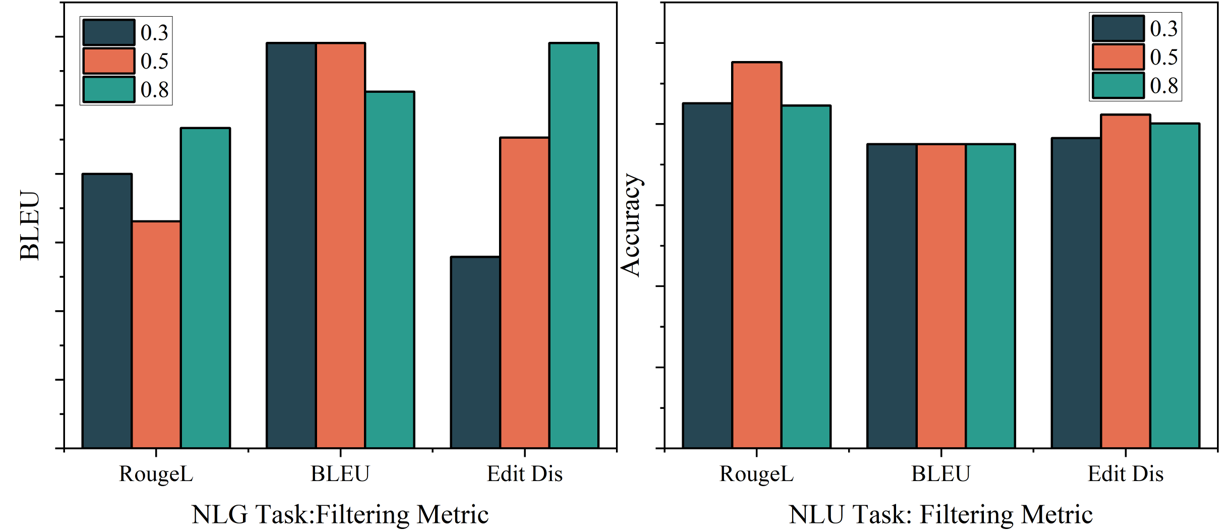}
  \caption{ 
  Results under different filtering metrics. We use three thresholds: $0.3$, $0.5$ and $0.8$.
  }
  \label{filtering}
\end{figure}

\subsection{Ablation Studies}

Figure~\ref{filtering} left shows the effects of using different filtering evaluation metrics at various thresholds for NLG tasks, with experiments conducted on the IT translation dataset. We find that using BLEU as a filtering metric in translation tasks yields stable results across thresholds of $0.3$, $0.5$, and $0.8$. Additionally, Edit Distance improves with higher thresholds, achieving better results than BLEU at a threshold of $0.8$. RougeL shows insensitivity in NLG tasks, failing to reach an acceptable level regardless of the threshold used. Its performance is unstable across thresholds, lacking a consistent trend in variation. 

\begin{table}[hbp]
\centering

\vspace{-5pt}

\scalebox{1}{
\begin{tabular}{cccccc} \toprule
           & Params & \multicolumn{2}{c}{BLEU} & \multicolumn{2}{c}{Edit Rate}  
           \\ \hline
           &        & IT         & Medical     & IT             & Medical          \\
Origin     & -      & 27.75      & 31.67       & 0.62           & 0.63
           \\
Mbart      & 406M   & 28.06      & 34.57       & 0.61           & 0.64             \\
Tiny-Mbart & 60M    & 26.23      & 29.36       & 0.55           & 0.58             \\
mT5        & 220M   & 26.9       & 32.25       & 0.61           & 0.63             \\
           &        & Law        & Koran       & Law            & Koran            \\
Origin     &  -     & 24.21      & 12.37       & 0.63           & 0.57 
          \\
Mbart      & 406M   & 26.94      & 12.51       & 0.63           & 0.56             \\
Tiny-Mbart & 60M    & 23.32      & 11.20        & 0.55           & 0.48             \\
mT5        & 220M   & 26.83      & 11.90        & 0.62           & 0.56           \\ \bottomrule 
\end{tabular}
}
\label{4}
\caption{
The experimental results with mbart tiny-mbart and mT5 rewriting models.
}
\end{table}

We also conduct comparative experiments on different filtering metrics in NLU tasks. The results, shown in Figure~\ref{filtering} right, illustrate the comparative outcomes for the sentiment analysis task. We find that using BLEU as a filtering metric in NLU tasks yields the same results across thresholds of $0.3$, $0.5$, and $0.8$, with minimal impact on outcomes. Additionally, Edit Distance remains relatively stable, with threshold variations causing minimal performance fluctuations. In NLU tasks, RougeL is more sensitive, achieving optimal results at a threshold of $0.5$.

We compare the performance of rewriting models across different architectures and the experimental results in the translation task are presented in Table~4. We select three language models with different parameter sizes and pre-training data. the results show that the large parameter rewrite model outperforms the small parameter rewrite model significantly. On the four datasets, the performance of using mbart-cc-25 is significantly higher than that of using tiny-mbart. We analyze that the large parameter model has a more abundant pre-training dataset and stronger understanding ability. In addition, when the parameter sizes are comparable, the rewrite performance of mbart is better than that of mT5. It can be seen that using mbart performs significantly better than using mT5 on the four domains.

\begin{figure}[hbp]
  \centering 
    \includegraphics[width=0.90\textwidth]{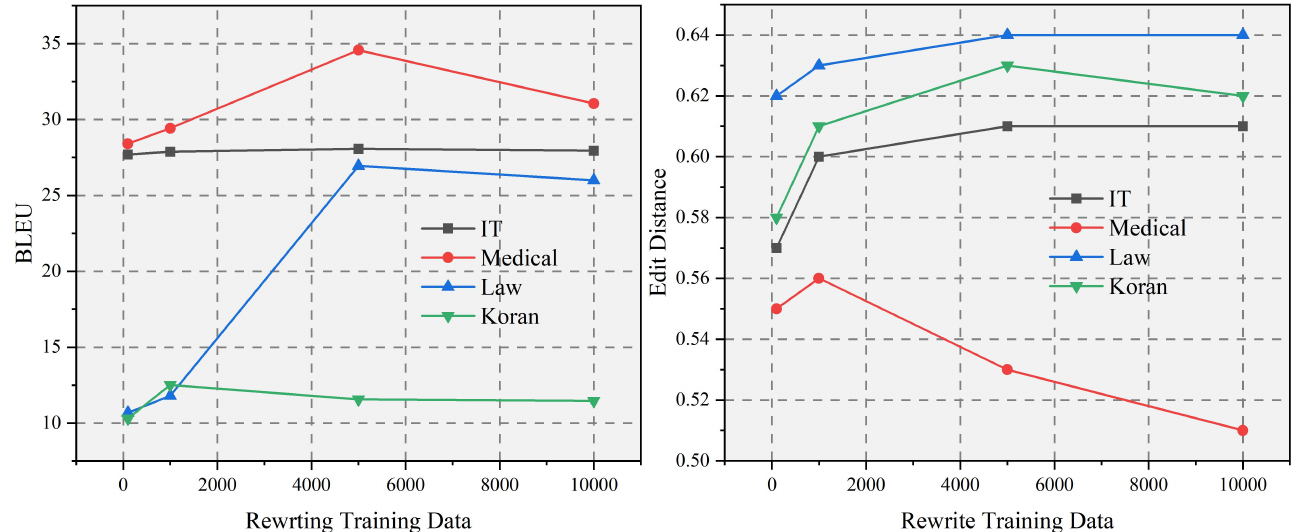}
  \caption{ 
  Translation performance when training a rewriting model using 100, 1000, 5000, and 10000 pieces of rewritten data, respectively.
  }
  \label{ablation_data_num}
  \vspace{-3mm}
\end{figure}

To systematically evaluate the practical utility of our ROI framework, we perform extensive experiments on multi-domain translation datasets by progressively scaling the size of training data for the rewriting model. Our experimental protocol initiates with a minimal set of randomly sampled data processed through ROI, followed by incremental expansions of the training corpus. As shown in Figure~\ref{ablation_data_num}, we observe a positive correlation between training data volume and both rewriting quality and downstream translation performance. Notably, it achieves superior translation performance compared to using original inputs across multiple domains with only approximately 5,000 instances of rewritten data. This finding suggests that the rewriting model can effectively capture appropriate reformulation patterns from a relatively small curated dataset. However, our analysis reveals a performance plateau when scaling beyond 10,000 rewritten instances, attributable to quality inconsistency. This observation provides empirical validation for the necessity of our proposed filtering algorithm in maintaining high-quality training instances. The non-monotonic improvement pattern further confirms that indiscriminate expansion of training data without quality control may introduce noise that diminishes model effectiveness.

\vspace{-5pt}

\section{Conclusion}
In this paper, we propose original input rewriting with filtering, a simple and versatile framework for optimizing input components to LLMs. This method mainly focuses on tasks in that the instruction component is relatively simple but the input part is important. We optimize the input component by rewriting model to make it more consistant with the preferences of LLMs for data. Through extensive experiments on multiple NLU and NLG datasets, we validate its effectiveness. The simplicity and efficacy of our framework make it a promising approach with substantial potential. 

\textbf{Limitations.}  Our proposed method has demonstrated promising performance on various versions of the Alpaca model. However, we acknowledge that we have not yet conducted experiments on larger-scale LLMs like GPT-3.5. At the same time, our method is primarily limited to single-turn question-answering language tasks. We look forward to addressing these issues in the future.

\bibliographystyle{colm2025_conference.bst}
\bibliography{colm2025_conference}


\end{document}